\documentclass[a4paper]{article}
\usepackage[dvips]{graphicx}
\usepackage{epsfig}
\usepackage{epsf}
\usepackage{subfigure}
\usepackage{amsmath}
\usepackage{amsopn}
\usepackage{amssymb}
\usepackage{amsfonts}
\usepackage{txfonts}
\usepackage{url}

\newtheorem{definition}{Definition}

\begin{document}

\title{A Novel Clustering Algorithm Based Upon Games on Evolving Network}
\author{Qiang Li$^{1}$\footnote{anjuh@cqu.edu.cn}, Zhuo Chen$^{2}$\footnote{czcq@sjtu.edu.cn}, Yan He$^{1}$\footnote{heyan@zju.edu.cn}, Jing-ping Jiang$^{1}$\footnote{eejiang@zju.edu.cn}\\
\\ \small
1 College of Electrical Engineering, Zhejiang University,\\ \small Hang Zhou, Zhejiang, 310027, China\\
\small 2 Department of Automation, Shanghai Jiao Tong University,\\
\small Shanghai, 200240, China} \maketitle

\begin{abstract}
This paper introduces a model based upon games on an evolving network, and
develops three clustering algorithms according to it. In the
clustering algorithms, data points for clustering are regarded as
players who can make decisions in games. On the network describing
relationships among data points, an edge-removing-and-rewiring (ERR)
function is employed to explore in a neighborhood of a data point,
which removes edges connecting to neighbors with small payoffs, and
creates new edges to neighbors with larger payoffs. As such, the
connections among data points vary over time. During the evolution
of network, some strategies are spread in the network. As a
consequence, clusters are formed automatically, in which data points
with the same evolutionarily stable strategy are collected as a
cluster, so the number of evolutionarily stable strategies
indicates the number of clusters. Moreover, the experimental results
have demonstrated that data points in datasets are clustered
reasonably and efficiently, and the comparison with other algorithms
also provides an indication of the effectiveness of the proposed
algorithms.
\end{abstract}

\maketitle

\section{Introduction}
Cluster analysis is an important branch of Pattern Recognition,
which is widely used in many fields such as pattern analysis,
data mining, information retrieval and image segmentation. For the
past thirty years, many excellent clustering algorithms have been
presented, say, \textit{K}-means \cite{MacQueen1967},
C4.5~\cite{Quinlan1993}, support vector clustering (SVC)
\cite{Ben-Hur2001}, spectral clustering~\cite{Ng2002}, etc., in
which the data points for clustering are fixed, and various
functions are designed to find separating hyperplanes. In recent
years, however, a significant change has been made. Some researchers
thought about that why not those data points could move by
themselves, just like agents or something, and collect together
automatically. Therefore, following their ideas, they created a few
exciting algorithms
\cite{Rhouma2001,Folino2002,Merwe2003,Labroche2003,Cui2006}, in
which data points move in space according to certain simple local
rules preset in advance.

Game theory came into being with the book named "Theory of Games and
Economic Behavior" by John von Neumann and Oskar Morgenstern
\cite{Neumann1944} in 1940. In this period, Cooperative Game was
widely studied. Till 1950's, John Nash published two well-known
papers to present the theory of non-cooperative game, in which he
proposed the concept of Nash equilibrium, and proved the existence
of equilibrium in a finite non-cooperative game
\cite{Nash1950,Nash1951}. Although non-cooperative game was
established on the rigorous mathematics, it required that players in
a game must be perfect rational or even hyper-rational. If this
assumption could not hold, the Nash equilibrium might not be reached
sometimes. On the other hand, evolutionary game theory
\cite{Smith1976} stems from the researches in biology which are to
analyze the conflict and cooperation between animals or plants. It
differs from classical game theory by focusing on the dynamics of
strategy change more than the properties of strategy equilibria, and
does not require perfect rational players. Besides, an important
concept, evolutionarily stable strategy \cite{Smith1976,Smith1973},
in evolutionary game theory was defined and introduced by John
Maynard Smith and George R. Price in 1973, which was often used to
explain the evolution of social behavior in animals.

To the best of our knowledge, the problem of data clustering has not
been investigated based on evolutionary game theory. So, if data
points in a dataset are considered as players in games, could
clusters be formed automatically by playing games among them? This
is the question that we attempt to answer. In our clustering
algorithm, each player hopes to maximize his own payoff, so he
constantly adjusts his strategies by observing neighbors' payoffs.
In the course of strategies evolving, some strategies are spread in
the network of players. Finally, some parts will be formed
automatically in each of which the same strategy is used. According
to different strategies played, data points in the dataset can be
naturally collected as several different clusters. The remainder of this
paper is organized as follows: Section 2 introduces some basic
concepts and methods about the evolutionary game theory and
evolutionary game on graph. In Section 3, the model based upon games
on evolving network is proposed and described specifically. Section
4 gives three algorithms based on this model, and the algorithms are
elaborated and analyzed in detail. Section 5 introduces those datasets used in the experiments briefly, and then demonstrates experimental results of the algorithms. Further, the relationship between the number of clusters and the number of nearest neighbors is discussed, and three edge-removing-and-rewiring (ERR) functions employed in the clustering algorithms are compared. The conclusion is given in Section 6.

\section{Related work}
Cooperation is commonly observed in genomes, cells, multi-cellular
organisms, social insects, and human society, but Darwin's Theory of
Evolution implies fierce competition for existence among selfish and
unrelated individuals. In past decades, many efforts have been
devoted to understanding the mechanisms behind the emergence and
maintenance of cooperation in the context of evolutionary game
theory.

Evolutionary game theory, which combines the traditional game theory
with the idea of evolution, is based on the assumption of bounded
rationality. On the contrary, in classical game theory players are
supposed to be perfectly rational or hyper-rational, and always
choose optimal strategies in complex environments. Finite
information and cognitive limitations, however, often make rational
decisions inaccessible. Besides, perfect rationality may cause the
so-called backward induction paradox \cite{Pettit1989} in finitely
repeated games. On the other hand, as the relaxation of perfect
rationality in classical game theory, bounded rationality means
people in games need only part rationality \cite{Simon1996}, which
explains why in many cases people respond or play instinctively
according to heuristic rules and social norms rather than adopting
the strategies indicated by rational game theory \cite{Szabo2007}.
So, various dynamic rules can be defined to characterize the
boundedly rational behavior of players in evolutionary game theory.

Evolutionary stability is a central concept in evolutionary game
theory. In biological situations the evolutionary stability provides
a robust criterion for strategies against natural selection.
Furthermore, it also means that any small group of individuals who
tries some alternative strategies gets lower payoffs than those who
stick to the original strategy \cite{Weibull1995}. Suppose that
individuals in an infinite and homogenous population who play
symmetric games with equal probability are randomly matched and all
employ the same strategy $A$. Nevertheless, if a small group of
mutants with population share $\epsilon\in(0,1)$ who plays some
other strategy appear in the whole group of individuals, they will
receive lower payoffs. Therefore, the strategy $A$ is said to be
evolutionary stable for any mutant strategy $B$, if and only if the
inequality, $E(A,(1-\epsilon)A+\epsilon
B)>E(B,(1-\epsilon)A+\epsilon B)$, holds, where the function
$E(\cdot,\cdot)$ denotes the payoff for playing strategy $A$ against
strategy $B$ \cite{Broom2000}.

In addition, the cooperation mechanism and spatial-temporal dynamics
related to it have long been investigated within the framework of
evolutionary game theory based on the prisoner's dilemma (PD) game
or snowdrift game which models interactions between a pair of
players. In early days, the iterated PD game was widely studied, in
which a player interacted with all other players. By round robin
interactions among players, strategies in the population began to
evolve according to their payoffs. As a result, the strategy of
unconditional defection was always evolutionary stable
\cite{Hofbauer1998} while pure cooperators could not survive.
Nevertheless, the Tit-for-Tat strategy is evolutionary stable as
well, which promotes cooperation based on reciprocity
\cite{Axelrod1981}.

Recently, evolutionary dynamics in structured populations has
attracted much attention, where the structured population denotes an infinite and well-mixed
population which simplifies the analytical description of the
evolution process. In real populations, individuals are more likely
affected by their neighbors than those who are far away, but the
spatial structure of population is omitted in the iterated PD game.
To study the spatial effects upon strategy frequencies in the
population, Nowak and May \cite{Nowak1993} have introduced the
spatial PD game, in which players are located on the vertices of a
two-dimensional lattice, whose edges represent connections among the
corresponding players. Instead of playing with all other
contestants, each player only interacts with his neighbors. Without
any strategic complexity the stable coexistence of cooperators and
defectors can be achieved. However, the model presented in
\cite{Nowak1993} assumes a noise free environment. To characterize
the effect of noise, Szab\'{o} and Toke \cite{Szabo1998} have
presented a stochastic update rule that permits irrational
choice. Besides, Perc and Szolnoki \cite{Perc2008} account for
social diversity by stochastic variables that determine the mapping
of game payoffs to individual fitness. Furthermore, many other works
centered on the lattice structure have also been done. For example,
Vukov and Szab\'{o} \cite{Vukov2005} have presented a hierarchical
lattice and shown that for different hierarchical levels the highest
frequency of cooperators may occur at the top or middle layer. For
more details about evolutionary games on graphs, see
\cite{Szabo2007,Doebeli2005,Nowak2006} and references therein.

Yet, as imitations of real social networks, the evolutionary game on
lattices assumes that there is a fixed neighborhood for each player.
Nevertheless, this assumption does not always hold for most of real
social networks. Unlike models mentioned above, the relationships among
players (data points) in our model are represented by a weighted and
directed network, which means that players are not located on a
regular lattice any more. And the network will evolve over time
because each player is allowed to apply an
edges-removing-and-rewiring (ERR) function to change his connections
between him and his neighbors. Furthermore, the payoff matrix of any
two players in the proposed model is also time-varying instead of a
constant payoff matrix, for instance, the payoff matrix in PD game.
As a consequence, when the evolutionarily stable strategies emerge
in the network, it will be observed that only a few players (data
points) receive considerable connections, while most of them have
only one connection. Naturally, players (data points) are divided
into several parts (clusters) according to their evolutionarily
stable strategies.

\section{Proposed model}
Assume a set $\textbf{\textit{X}}$ with $N$ players,
$\textbf{\textit{X}}=\{\textbf{\textit{X}}_{1},\textbf{\textit{X}}_{2},\cdots,\textbf{\textit{X}}_{N}\}$,
which are distributed in a \textit{m}-dimensional metric space. In
this metric space, there is a distance function
$d:\textbf{\textit{X}} \times \textbf{\textit{X}} \longrightarrow
\mathbb{R}$, which satisfies the condition that the closer any two
players are, the smaller the output is. Based on the distance
function a distance matrix is computed whose entries are distances
between any two players. Next, a weighted and directed
\textit{k}-nearest neighbor (knn) network,
$G_{0}(\textbf{\textit{X}},E_{0},d)$, is formed by adding \textit{k}
edges directed toward its \textit{k} nearest neighbors for each
player, which represents the initial relationships among all players.
\begin{definition}
If there is a set $\textbf{\textit{X}}$ with $N$ players,
$\textbf{\textit{X}}=\{\textbf{\textit{X}}_{1},\textbf{\textit{X}}_{2},\cdots,\textbf{\textit{X}}_{N}\}$,
the initial weighted and directed knn network,
$G_{0}(\textbf{\textit{X}},E_{0},d)$, is created as below.
\begin{equation}
\left\{\begin{array}{ll}\textbf{\textit{X}}=\big\{\textbf{\textit{X}}_{i},i=1,2,\cdots,N\big\}\vspace{4pt}\\
E_{0}=\bigcup^{N}_{i=1}E_{0}(i)\vspace{4pt}\\
E_{0}(i)=\Big\{e_{0}\big(\textbf{\textit{X}}_{i},\textbf{\textit{X}}_{j}\big)\mid
j\in\Gamma_{0}(i)\Big\}\vspace{4pt}\\
\Gamma_{0}(i)=\bigg\{j \Big| j=\underset{\textbf{\textit{X}}_{h}\in
\textbf{\textit{X}}}{argmink}\bigg(\Big\{d(\textbf{\textit{X}}_{i},\textbf{\textit{X}}_{h}),\textbf{\textit{X}}_{h}\in\textbf{\textit{X}}\Big\}\bigg)\bigg\}
\end{array} \right.
\end{equation}
Here, players in the set $\textbf{\textit{X}}$ correspond to
vertexes in the network $G_{0}(\textbf{\textit{X}},E_{0},d)$;
directed edges in the network represent certain relationships
established among players and the distances denote the weights over
edges; the function, $argmink(\cdot)$, is to find \textit{k} nearest
neighbors of a player, which construct a neighbor set,
$\Gamma_{0}(i)$.
\end{definition}

It is worth noting that the distance between a player
$\textbf{\textit{X}}_{i}$ and himself,
$d(\textbf{\textit{X}}_{i},\textbf{\textit{X}}_{i})$, is zero
according to the defined distance function, which means that at the beginning he is
one of his \textit{k} nearest neighbors. So there is an edge between
the player $\textbf{\textit{X}}_{i}$ and himself, namely a
self-loop. In practice, the distance is set by
$d(\textbf{\textit{X}}_{i},\textbf{\textit{X}}_{i})=1$.

When the initial network $G_{0}$ is established, we can define a
evolutionary game,
$\Omega=\{\textbf{\textit{X}},G_{0},S_{0},U_{0}\}$, on it further.
\begin{definition}
An evolutionary game
$\Omega=\{\textbf{\textit{X}},G_{0},S_{0},U_{0}\}$ on a network
$G_{0}$ is a 4-tuple: $\textbf{\textit{X}}$ is a set of players;
$G_{0}$ represents the initial relationships among players;
$S_{0}=\{s_{0}(i),i=1,2,\cdots,N\}$ represents a set of players'
strategies; $U_{0}=\{u_{0}(i),i=1,2,\cdots,N\}$ represents a set of
players' payoffs. In each round, players choose theirs strategies
simultaneously, and each player can only observe its neighbors'
payoffs, but does not know the strategy profile of anyone of all
other players in $\textbf{\textit{X}}$. Finally, all players update
their strategy profiles synchronously.
\end{definition}

In the proposed model, assume each player in $\textbf{\textit{X}}$
sets up a group, and hopes to maximize the payoff of his own group
in order to attract more players to join. At the same time, he also
joins \textit{k} groups set up by other players, so the initial
strategy set $s_{0}(i)\in S_{0}$ of a player
$\textbf{\textit{X}}_{i}$ is defined as his neighbor set,
$s_{0}(i)=\Gamma_{0}(i)$. However, it is worth noting that his
preference to join each group is changeable, whose initial value is
given below,
\begin{equation}
\begin{array}{ll}
P_{0}(i)=\Big\{p_{0}(i,j),j\in\Gamma_{0}(i)\Big\}\vspace{4pt}\\
p_{0}(i,j)=1/\big|\Gamma_{0}(i)\big|=1/k
\end{array}
\end{equation}
where $P_{0}(i)$ is the preference set, and the symbol $|\cdot|$
denotes the cardinality of a set. Thus, a player's payoff may be
defined as follows.

\begin{definition}
After a player $\textbf{\textit{X}}_{i}$ chooses his strategies and
corresponding preferences, he receives a payoff $u_{0}(i)$,
\begin{equation}
\begin{array}{ll}
u_{0}(i)=\sum_{j\in\Gamma_{0}(i)}R(i,j)\vspace{4pt}\\
R(i,j)=p_{0}(i,j)\times
Deg_{0}(j)/d(\textbf{\textit{X}}_{i},\textbf{\textit{X}}_{j})
\end{array}
\end{equation}
where $Deg_{0}(j)$ represents the degree of a player
$\textbf{\textit{X}}_{j}$ in the neighbor set, and the degree is a sum of the
indegree and outdegree.
\end{definition}

When all players have received their payoffs, each one will check
his neighbors' payoffs, and apply an ERR function $B_{i}(\cdot)$ to
change his connections and update his neighbor set.

\begin{definition}
The ERR function $B_{i}(\cdot)$ is a function of payoffs, whose
output is a set with \textit{k} elements, i.e., an updated neighbor
set $\Gamma_{1}(i)$ of a player $\textbf{\textit{X}}_{i}$.
\begin{equation}
\begin{array}{lll}
\Gamma_{1}(i)=B_{i}\big(\widehat{u}_{0}(i)\big)=\underset{j\in\Gamma_{0}(i)\bigcup\Upsilon_{0}(i)}{argmaxk}\Big(\big\{u_{0}(j),j\in\Gamma_{0}(i)\bigcup\Upsilon_{0}(i)\big\}\Big)\vspace{4pt}\\
\widehat{u}_{0}(i)=\Big\{u_{0}(j),j\in\Gamma_{0}(i)\bigcup\Upsilon_{0}(i)\Big\},\Upsilon_{0}(i)=\bigcup_{j\in\Gamma^{+}_{0}(i)}\Gamma_{0}(j)\vspace{4pt}\\
\Gamma^{+}_{0}(i)=\Big\{j|u_{0}(j)\geq\theta_{0}(i),j\in\Gamma_{0}(i)\Big\},\Gamma^{-}_{0}(i)=\Gamma_{0}(i)\backslash\Gamma^{+}_{0}(i)
\end{array}
\end{equation}
where $\theta_{0}(i)$ is a payoff threshold, $\Upsilon_{0}(i)$ is
called an extended neighbor set, and the function $argmaxk(\cdot)$
is to find \textit{k} neighbors with the first to the \textit{k}-th
largest payoffs in the union $\Gamma_{0}(i)\bigcup\Upsilon_{0}(i)$.
\end{definition}

The ERR function $B_{i}(\cdot)$ expands the view of a player
$\textbf{\textit{X}}_{i}$, and makes him able to observe payoffs of
players in the extended neighbor set, which provides a chance to
find players with higher payoffs around him. If no players with
higher payoffs are found in the extended neighbor set, i.e.,
$min(\{u_{0}(j),j\in\Gamma_{0}(i)\})\geq
max(\{u_{0}(h),h\in\Upsilon_{0}(i)\})$, then the output of the ERR
function is $\Gamma_{1}(i)=B_{i}(\widehat{u}_{0}(i))=\Gamma_{0}(i)$.
Otherwise, a neighbor with the minimal payoff will be removed
together with the corresponding edge from the neighbor set and the edge
set, and replaced by a found player with larger payoff. This process
is repeated till the payoffs of unconnected players in the extended
neighbor set are no larger than those of connected neighbors. Since
the connections among players, namely the edge set $E_{0}$ in the
network $G_{0}(\textbf{\textit{X}},E_{0},d)$, are changed by the ERR
function, the network  $G_{t}(\textbf{\textit{X}},E_{t},d)$ will
begin to evolve over time, when  $t\geq1$. As such, after the ERR
function is applied, the new preference set $P_{t}(i)$ of a player
$\textbf{\textit{X}}_{i}$ needs to be adjusted.

\begin{definition}
The new preference set of a player
$\textbf{\textit{X}}_{i}\in\textbf{\textit{X}}$ is formed by means
of the below formulation.
\begin{equation}
\begin{array}{lll}
P_{t}(i)=\Big\{p_{t}(i,j),j\in\Gamma_{t}(i)\Big\}\vspace{4pt}\\
p_{t}(i,j)=\left\{\begin{array}{ll}
\frac{\sum_{h\in(\Gamma_{t-1}(i)\backslash\Delta)} p_{t-1}(i,h)}
{\big|\Gamma_{t}(i)\backslash\Delta\big|}
& \textrm{ if $j\in\Gamma_{t}(i)\backslash\Delta$}\vspace{4pt}\\
p_{t-1}(i,j) & \textrm{ otherwise}
\end{array} \right.\\
\Delta=\big\{\Gamma_{t-1}(i)\bigcap\Gamma_{t}(i)\big\}
\end{array}
\end{equation}
Then, the player adjusts his preference set $P_{t}(i)$ as follows.
First, he identifies the neighbor $\textbf{\textit{X}}_{m}$ with maximal
payoff in his neighbor set,
\begin{equation}
m=\underset{j\in\Gamma_{t}(i)}{argmax}\Big(\big\{u_{t-1}(j),j\in\Gamma_{t}(i)\big\}\Big)
\end{equation}
Next, each element in the preference set is taken its square root
and the preference $p_{t}(i,m)$ of joining the group built by the
neighbor $\textbf{\textit{X}}_{m}$ becomes negative,
\begin{equation}
\left\{\begin{array}{ll}P_{t}(i)=\big\{\sqrt{p_{t}(i,j)},j\in\Gamma_{t}(i)\big\}\vspace{4pt}\\
\sqrt{p_{t}(i,m)}=-\sqrt{p_{t}(i,m)},m\in\Gamma_{t}(i)
\end{array} \right.
\end{equation}
Further, let
$Ave_{t}(i)=(\sum_{j\in\Gamma_{t}(i)}\sqrt{p_{t}(i,j)})/|\Gamma_{t}(i)|$,
thus, the updated preference set is,
\begin{equation}
\begin{array}{ll}
P_{t}(i)=\big\{p_{t}(i,j),j\in\Gamma_{t}(i)\big\}\vspace{4pt}\\
p_{t}(i,j)=\Big(2\times Ave_{t}(i)-\sqrt{p_{t}(i,j)}\Big)^{2}
\end{array}
\end{equation}
\end{definition}

After the preference set of each player has been adjusted, an
iteration of the model is completed. In conclusion, when $t\geq1$,
the network representing relationships among players begins to evolve
over time, which also makes a player's strategy set and payoff set
become time-varying. Therefore, the game on evolving network
$G_{t}(\textbf{\textit{X}},E_{t},d)$ is rewritten as
$\Omega=\{\textbf{\textit{X}},G_{t},S_{t},U_{t}\}$.
\begin{equation}
\begin{array}{lll}
G_{t}(\textbf{\textit{X}},E_{t},d)=\left\{\begin{array}{ll}
\textbf{\textit{X}}(t)=\Big\{\textbf{\textit{X}}_{i}(t),i=1,2,\cdots,N\Big\}\vspace{4pt}\\
\Gamma_{t}(i)=B_{i}(\widehat{u}_{t-1}(i))\vspace{4pt}\\
E_{t}=\bigcup^{N}_{i=1}E_{t}(i)\vspace{4pt}\\
E_{t}(i)=\Big\{e_{t}\big(\textbf{\textit{X}}_{i},\textbf{\textit{X}}_{j}\big)\mid
j\in\Gamma_{t}(i)\Big\}\end{array} \right.\vspace{4pt}\\
S_{t}=\Big\{s_{t}(i)\big|s_{t}(i)=\Gamma_{t}(i),i=1,2,\cdots,N\Big\}\vspace{4pt}\\
U_{t}=\Big\{u_{t}(i)\big|u_{t}(i)=\sum_{j\in\Gamma_{t}(i)}p_{t}(i,j)\times
Deg_{t}(j)/d(\textbf{\textit{X}}_{i},\textbf{\textit{X}}_{j}),i=1,2,\cdots,N\Big\}
\end{array}
\end{equation}

As the model is iterated, some strategies are spread in the evolving
network, which are played by a great number of players. In other
words, a certain strategy or several strategies in the strategy set
$s_{t}(i)\in S_{t}$ will be always played by the player
$\textbf{\textit{X}}_{i}$ with the maximal preference.
\begin{definition}
If a player $\textbf{\textit{X}}_{i}\in\textbf{\textit{X}}$ always
or periodically chooses a strategy $\widehat{s}_{t}(i,j)\in
s_{t}(i)$ with the maximal preference
$max(\{p_{t}(i,j),j\in\Gamma_{t}(i)\})$,
\begin{equation}
\left\{\begin{array}{ll}
\widehat{s}_{t}(i,j)=\underset{j\in\Gamma_{t}(i)}{argmax}\big\{p_{t}(i,j),j\in\Gamma_{t}(i)\big\}\vspace{4pt}\\
\widehat{s}_{t}(i,j)=\widehat{s}_{t-1}(i,j)\vspace{4pt}\\
\widehat{s}_{t}(i,j)=\widehat{s}_{t-nT}(i,j)
\end{array} \right.
\end{equation}
then the strategy $\widehat{s}_{t}(i,j)\in s_{t}(i)$ is called the
\emph{evolutionarily stable strategy (ESS)} of the player
$\textbf{\textit{X}}_{i}\in\textbf{\textit{X}}$. Here, the variable
$T$ is a constant period.
\end{definition}

As a consequence, each player in the network will choose one of
evolutionarily stable strategies as his strategy and he is not willing
to change his strategy unilaterally during the iterations.

\section{Algorithm and analysis}
In this section, at first three different ERR functions
($B^{1}_{i}(\cdot),B^{2}_{i}(\cdot),B^{3}_{i}(\cdot)$) are designed,
and then three clustering algorithms (EG1, EG2, EG3) based on them
are established. Finally, the clustering algorithm is elaborated and
analyzed in detail.

\subsection{Clustering algorithms}
Assume an unlabeled dataset
$\textbf{\textit{X}}=\{\textbf{\textit{X}}_{1},\textbf{\textit{X}}_{2},\cdots,\textbf{\textit{X}}_{N}\}$,
in which each instance consists of $m$ features. In the clustering
algorithms, the relationships among all data points are represented by a
weighted and directed network $G_{t}(\textbf{\textit{X}},E_{t},d)$,
and each data point in the dataset is considered as a player in the
proposed model, who adjusts his strategy profile in order to
maximize his own payoff by observing other players' payoffs in the
union $\Gamma_{t}(i) \cup \Upsilon_{t}(i)$.

According to the proposed model, after a distance function
$d:\textbf{\textit{X}} \times \textbf{\textit{X}} \longrightarrow
\mathbb{R}$ is selected, the initial connections of data points,
$G_{0}(\textbf{\textit{X}},E_{0},d)$, are constructed by means of
Definition 1. Then, the initial payoff set $U_{0}$ of data points is
computed step by step. Finally, an ERR function is applied to
explore in a neighborhood of each data point, which changes
neighbors in the neighbor set of the data point. Thus, a network
$G_{t}(\textbf{\textit{X}},E_{t},d)$ will be evolving when $t\geq1$.

However, different ERR functions provide different exploring
capacities for data points, i.e., the observable areas of data
points depend on an ERR function. As such, there is no doubt that
the obtained results vary when different ERR functions are employed.
Here, three ERR functions
($B^{1}_{i}(\cdot),B^{2}_{i}(\cdot),B^{3}_{i}(\cdot)$) are designed,
and three clustering algorithms based on them are constructed
respectively.
\\

Algorithm EG1:
\\

In Algorithm EG1, an ERR function $B^{1}_{i}(\cdot)$ that is
realized most easily is used. This function always observes an
extended neighbor set formed by
$\lceil\eta\times|\Gamma_{t-1}(i)|\rceil$ neighbors of a
data point $\textbf{\textit{X}}_{i}$, where the symbol
$\lceil\cdot\rceil$ is to take an integer part of a number
satisfying the integer part is no larger than the number, and the
variable $\eta$ is called a ratio of exploration, $\eta\in[0,1]$.
According to Definition 4, a payoff threshold $\theta_{t-1}^{1}(i)$ is set
by
$\theta_{t-1}^{1}(i)=find^{\alpha}(\{u_{t-1}(j),j\in\Gamma_{t-1}(i)\}),\alpha=\lceil(1-\eta)\times|\Gamma_{t-1}(i)|\rceil$
firstly, where the function $find^{\alpha}(\cdot)$ is to find the
\textit{$\alpha$}-th largest payoff in the neighbor set
$\Gamma_{t-1}(i)$. As such, the set $\Gamma_{t-1}(i)$ is divided
into two sets naturally:
\begin{equation}
\Gamma^{+}_{t-1}(i)=\Big\{j\big|u_{t-1}(j)\geq\theta_{t-1}^{1}(i),j\in\Gamma_{t-1}(i)\Big\},\Gamma^{-}_{t-1}(i)=\Gamma_{t-1}(i)\backslash\Gamma^{+}_{t-1}(i)
\end{equation}

Then, based on the set $\Gamma^{+}_{t-1}(i)$, the extended neighbor
set is built,
$\Upsilon_{t-1}(i)=\bigcup_{j\in\Gamma^{+}_{t-1}(i)}\Gamma_{t-1}(j)$.
Further, the observable payoff set of the data point is written as,
\begin{equation}
\widehat{u}_{t-1}(i)=\Big\{u_{t-1}(j),j\in\Gamma_{t-1}(i)\cup\Upsilon_{t-1}(i)\Big\}
\end{equation}
At last, the ERR function $B^{1}_{i}(\cdot)$ is applied, which means
that the edges connecting to the neighbors with small payoffs are
removed and new edges are created between the data point and found
players with larger payoffs. Hence, his new neighbor set is
\begin{equation}
\Gamma_{t}(i)=B^{1}_{i}(\widehat{u}_{t-1}(i))=\underset{j\in\Gamma_{t-1}(i)\cup\Upsilon_{t-1}(i)}{argmaxk}\Big(\big\{u_{t-1}(j),j\in\Gamma_{t-1}(i)\cup\Upsilon_{t-1}(i)\big\}\Big)
\end{equation}
\\

Algorithm EG2:
\\

Unlike Algorithm EG1, the ERR function $B^{2}_{i}(\cdot)$ in
Algorithm EG2 adjusts the number of neighbors dynamically to form an
extended neighbor set instead of the constant number of neighbors in
Algorithm EG1. Furthermore, the payoff threshold
$\theta_{t-1}^{2}(i)$ is set by the average of neighbors' payoffs,
$\theta_{t-1}^{2}(i)=\sum_{j\in\Gamma_{t-1}(i)}u_{t-1}(j)/|\Gamma_{t-1}(i)|$.
Next, the set $\Gamma^{+}_{t-1}(i)$ is formed,
$\Gamma^{+}_{t-1}(i)=\{j|u_{t-1}(j)\geq\theta_{t-1}^{2}(i),j\in\Gamma_{t-1}(i)\}$,
and then the new neighbor set is achieved by means of the ERR
function $B^{2}_{i}(\cdot)$. In the case, when the payoffs of all
neighbors are equal to the payoff threshold $\theta_{t-1}^{2}(i)$,
the output of the ERR function is
$\Gamma_{t}(i)=B^{2}_{i}(\widehat{u}_{t-1}(i))=\Gamma_{t-1}(i)$.
This may be viewed as self-protective behavior for avoiding a payoff
loss due to no enough information acquired.
\\

Algorithm EG3:
\\

The ERR function $B^{3}_{i}(\cdot)$ used in Algorithm EG3 provides
more strongly exploring capacities for the data points with small
payoffs than that for those with larger payoffs. Generally speaking,
for maximizing their payoffs, players with small payoffs often seems
radical and show stronger desire for exploration, because this is
the only way to improve their payoffs. On the other hand, those
players with large payoffs look conservative for protecting their
payoff gotten.

Formally, the ratio of exploration $\gamma(i)$ of a data point
$\textbf{\textit{X}}_{i}\in\textbf{\textit{X}}$ is given as below:
\begin{equation}
\gamma(i)=\frac{(\underset{j\in\textbf{\textit{X}}}{max}(u_{t-1}(j))+\underset{j\in\textbf{\textit{X}}}{min}(u_{t-1}(j)))-u_{t-1}(i)}
{\underset{j\in\textbf{\textit{X}}}{max}\big(u_{t-1}(j)\big)}
\end{equation}
Thus, the data point $\textbf{\textit{X}}_{i}\in\textbf{\textit{X}}$
can observe an extended neighbor set that is formed by
$\lceil\gamma(i)\times|\Gamma_{t-1}(i)|\rceil$ neighbors.
Further, the payoff threshold is set by
$\theta_{t-1}^{3}(i)=find^{\beta(i)}(\{u_{t-1}(j),j\in\Gamma_{t-1}(i)\}),\beta(i)=\lceil(1-\gamma(i))\times|\Gamma_{t-1}(i)|\rceil$
and then the new neighbor set $\Gamma_{t}(i)$ is built according to
the ERR function $B^{3}_{i}(\cdot)$.

The ERR function brings about changes of the connections among data
points, so that the preferences need to be adjusted in terms of
Definition 5. When the evolutionarily stable strategies appear in the
network, the clustering algorithm exits. In the end, the data points
using the same evolutionarily stable strategy are gathered together
as a cluster, and the number of evolutionarily stable strategies
indicates the number of clusters.


\subsection{Analysis of algorithm}
The process of data clustering in the proposed algorithm can be
viewed an explanation about the group formation in society.
Initially, each data point (a player) in the dataset establishes a
group which corresponds to an initial cluster at the same time that
he joins other groups built by \textit{k} other players. As such,
the preference $p(i,j)$ may be explained as the level of
participation; $Deg(i)$ represents the total number of players in a
group; $1/d(\textbf{\textit{X}}_{i},\textbf{\textit{X}}_{j})$
denotes the position that the player occupies in a group in order to
identify a player is a president or an average member, and a player
$\textbf{\textit{X}}_{i}$ usually occupies the highest position in
his own group; the total payoff $u(i)$ of a player
$\textbf{\textit{X}}_{i}$ is viewed as the attraction of a group.
According to Definition 3, the reward of a player $\textbf{\textit{X}}_{i}$
is associated directly with the preference, the total number of
players and his position in a group. If a player
$\textbf{\textit{X}}_{i}$ who occupies an important position joins a
group with maximal preference, and the number of players in the
group is considerable, then the reward $R(i,j)$ that the player
receives is also large. On the other hand, for a player
$\textbf{\textit{X}}_{h}$ who is an average member in the same
group, i.e.,
$1/d(\textbf{\textit{X}}_{i},\textbf{\textit{X}}_{j})>1/d(\textbf{\textit{X}}_{h},\textbf{\textit{X}}_{j})$,
although his preference to join this group is as same as that of the
player $\textbf{\textit{X}}_{i}$, $p(i,j)=p(h,j)$, his reward
acquired from this group is smaller than that of the player
$\textbf{\textit{X}}_{i}$ by means of Eq.~3. This seems unfair, but
it is consistent with the phenomena observed in society. In
addition, the total payoff $u_{t}(i)$ of the player
$\textbf{\textit{X}}_{i}$ is the sum of his all rewards,
$u_{t}(i)=\sum_{j\in\Gamma_{t}(i)}R(i,j)$.

Certainly, each player is willing to join a group with large
attraction, and quit a group with little attraction, as is done
by an ERR function. Next, the player finds the group with the
largest attraction and increases the level of participation, namely
the preference $p_{t}(i,j)$ at the same time that other preferences
are decreased. To adjust the preference set $P_{t}(i)$ of a player
$\textbf{\textit{X}}_{i}$, the Grover iteration $G$ in the quantum
search algorithm \cite{Grover1997}, a well-known algorithm in
quantum computation\cite{Nielsen2000}, is employed, which is a way
to adjust the probability amplitude of each term in a superposition
state. By adjustment, the probability amplitude of the wanted is
increased, while the others are reduced. This whole process may be
regarded as the \emph{inversion about average} operation
\cite{Grover1997}. For our case, each element in the preference set
needs to be taken its square root first, and then the average
$Ave_{t}(i)$ of all square roots 
is obtained. Finally, all values are inverted about the average.
There are three main reasons that we select the modified Grover
iteration as the updating method of preferences: (a) the sum of
preferences updated retains one,
$\sum_{j\in\Gamma_{t}(i)}p_{t}(i,j)=1$, (b) a certain preference
updated in a player's preference set is much larger than the others,
$p_{t}(i,j)\gg p_{t}(i,h),h\in\Gamma_{t}(i)\backslash j$, and (c) it
helps players' payoffs to be a power law distribution, in which only
a few players' payoffs are far larger than others' after iterations.
Moreover, this is consistent with our observations in society, i.e.,
a player will change the level of participation obviously after he
takes part in activities held by neighbors' groups. In other
words, for the group with large attraction he shows higher level of
participation, whereas he hardly joins those groups with little
attraction.

After the model is iterated several times, a player
$\textbf{\textit{X}}_{i}\in\textbf{\textit{X}}$ will find the most
attractive group for him, and join this group with the largest
preference, $max(\{p_{t}(i,j),j\in\Gamma_{t}(i)\})$. Hence, only a
few groups are so attractive that almost all players join them. On
the other hand, most of groups are closed due to a lack of players.
Those lucky survivals not only attract many players but also those
players in the groups show the highest level of participation.
Furthermore, these surviving groups are the clusters formed by
players (data points) automatically, where the players (data points)
in a cluster play the same evolutionarily stable strategies, and the
number of surviving groups is also the number of clusters.

\section{Experiments and Discussions}
To evaluate these three clustering algorithms, five datasets are
selected from UCI repository \cite{Blake1998}, which are Soybean,
Iris, Wine, Ionosphere and Breast cancer Wisconsin datasets, and
experiments are performed on them.

\subsection{Experiments}
In this subsection, firstly these datasets are introduced briefly,
and then the experimental results are demonstrated.
The original data points in above datasets all are scattered in high
dimensional spaces spanned by their features, where the description
of all datasets is summarized in Table~\ref{tab:2}. As for Breast
dataset, some lost features are replaced by random numbers, and the Wine dataset is standardized. Finally,
this algorithm is coded in Matlab 6.5.
\begin{table}[htbp]
\caption{Description of datasets.}\label{tab:2} \centering
{\begin{tabular} {cccc} \hline Dataset & Instances & Features &
classes
\\\hline
Soybean & \hphantom{0}47 & \hphantom{0}21 & 4 \\
Iris & \hphantom{0}150 & \hphantom{0}4 & 3 \\
Wine & \hphantom{0}178 & \hphantom{0}13 & 3 \\
Ionosphere & \hphantom{0}351 & \hphantom{0}32 & 2\\
Breast & \hphantom{0}699 & \hphantom{0}9 & 2\\
\hline
\end{tabular}}
\end{table}

Throughout all experiments, data points in a dataset are considered
as players in games whose initial positions are taken from the
dataset. Next, the network representing initial relationships among data
points are created according to Definition 1, after a distance function is
selected. This distance function only needs to satisfy the condition
that the more similar data points are, the smaller the output of the
function is. In the experiments, the distance function applied is as
following:
\begin{equation}
d\big(\textbf{\textit{X}}_{i},\textbf{\textit{X}}_{j}\big)=exp\Big(\|\textbf{\textit{X}}_{i}-\textbf{\textit{X}}_{j}\|/2\sigma^{2}\Big),i,j=1,2,\cdots,N
\end{equation}
where the symbol $\|\cdot\|$ represents $L2$-norm. The advantage of
this function is that it not only satisfies above requirements, but
also overcomes the drawbacks of Euclidean distance. For instance,
when two points are very close, the output of Euclidean distance
function approaches zero, as may make the computation of payoff fail
due to the payoff approaching infinite. Nevertheless, when Eq.(15)
is selected as the distance function, it is more convenient to
compute the players' payoffs, since its minimum is one and the
reciprocals of its output are between zero and one,
$1/d(\textbf{\textit{X}}_{i},\textbf{\textit{X}}_{j})\in[0,1]$. In
addition, the parameter $\sigma$ in Eq.(15) takes one and the
distance between a data point and itself is set by
$d(\textbf{\textit{X}}_{i},\textbf{\textit{X}}_{i})=1$. As is
analyzed in 4.2, a data point $\textbf{\textit{X}}_{i}$ occupies the
highest position in the group established by himself.

\begin{figure}[htbp]
\centering \subfigure[Soybean dataset]{
\includegraphics[width=0.3\textwidth]{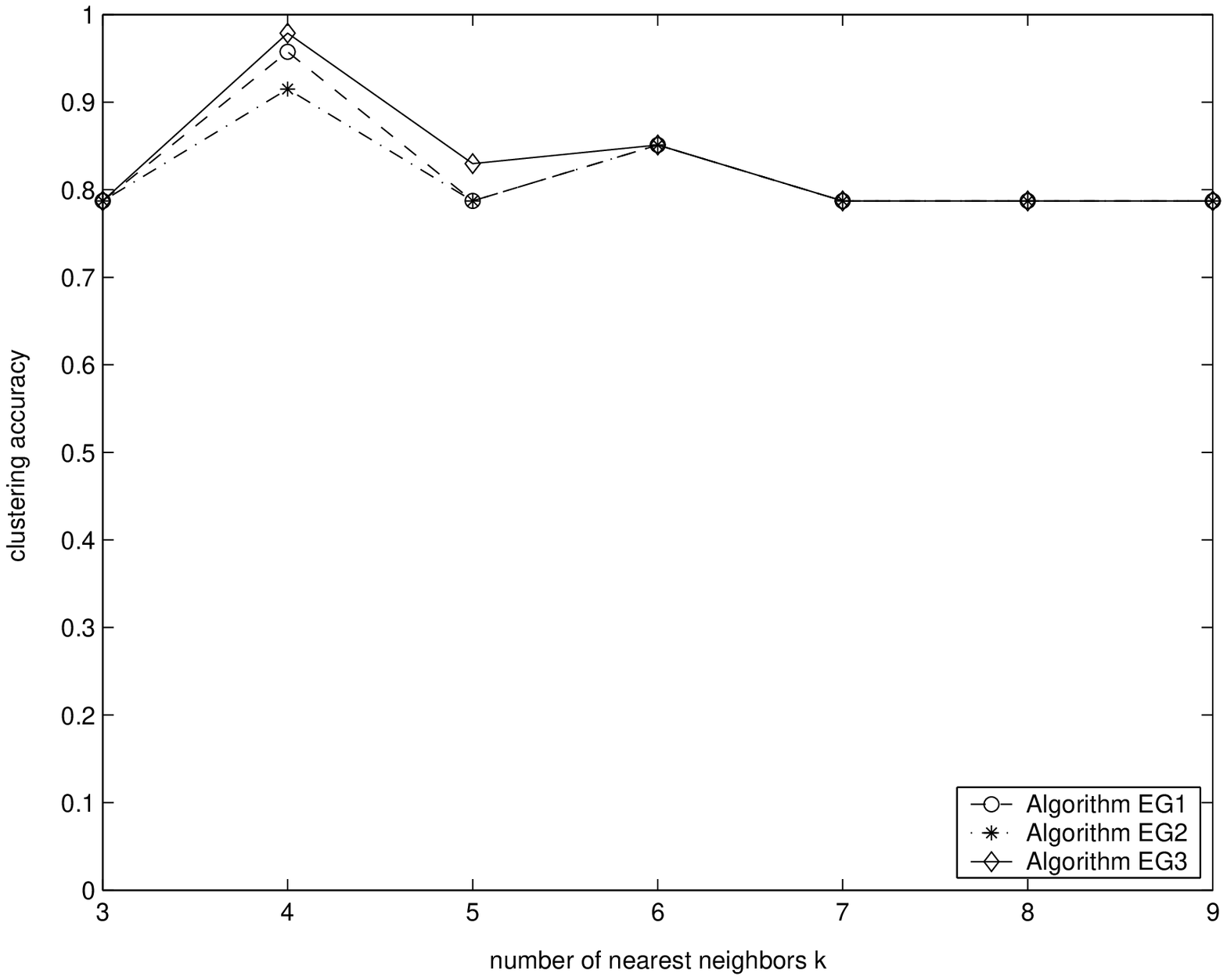}} \subfigure[Iris dataset]{
\includegraphics[width=0.3\textwidth] {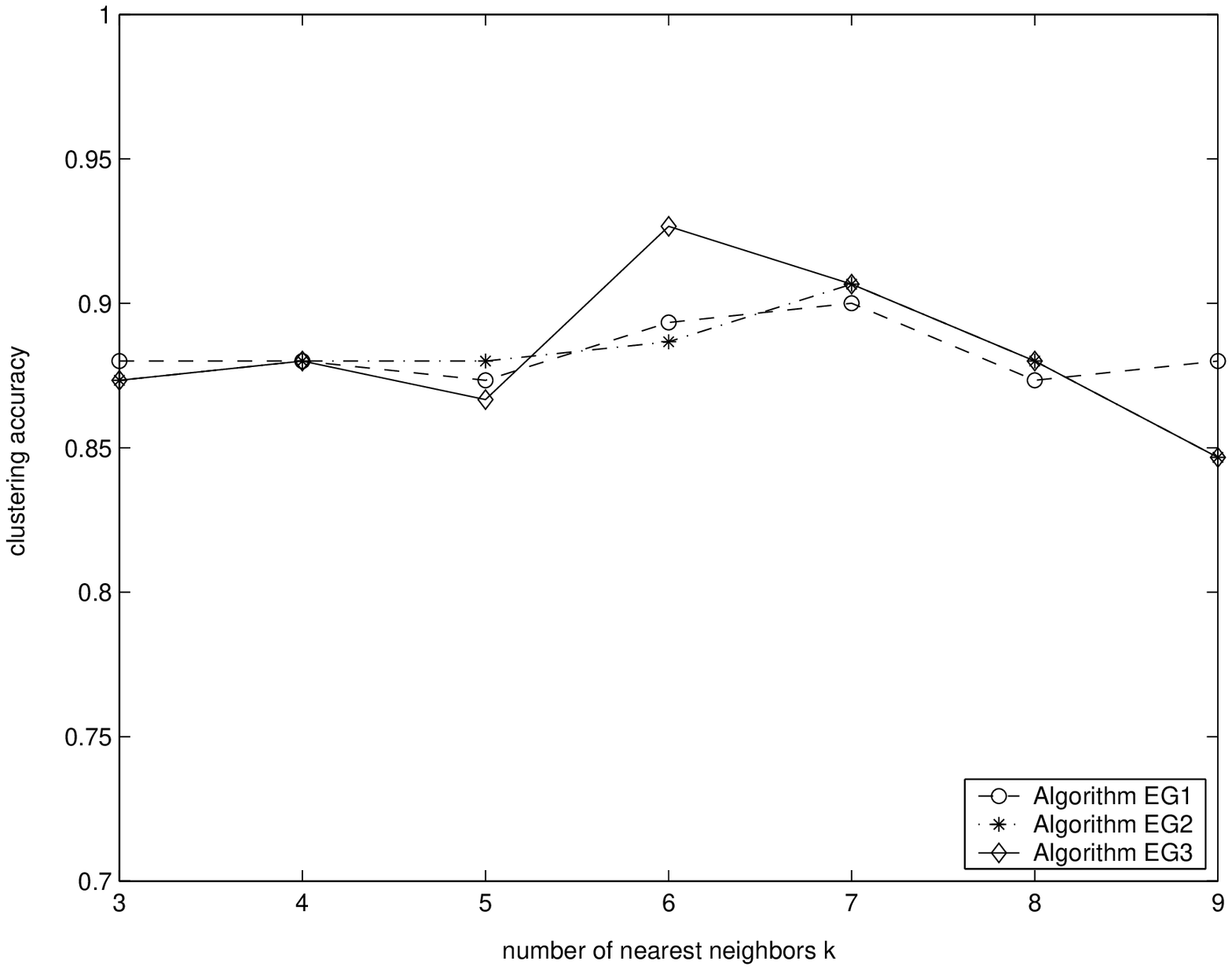}}
\subfigure[Wine dataset]{ \includegraphics[width=0.3\textwidth]{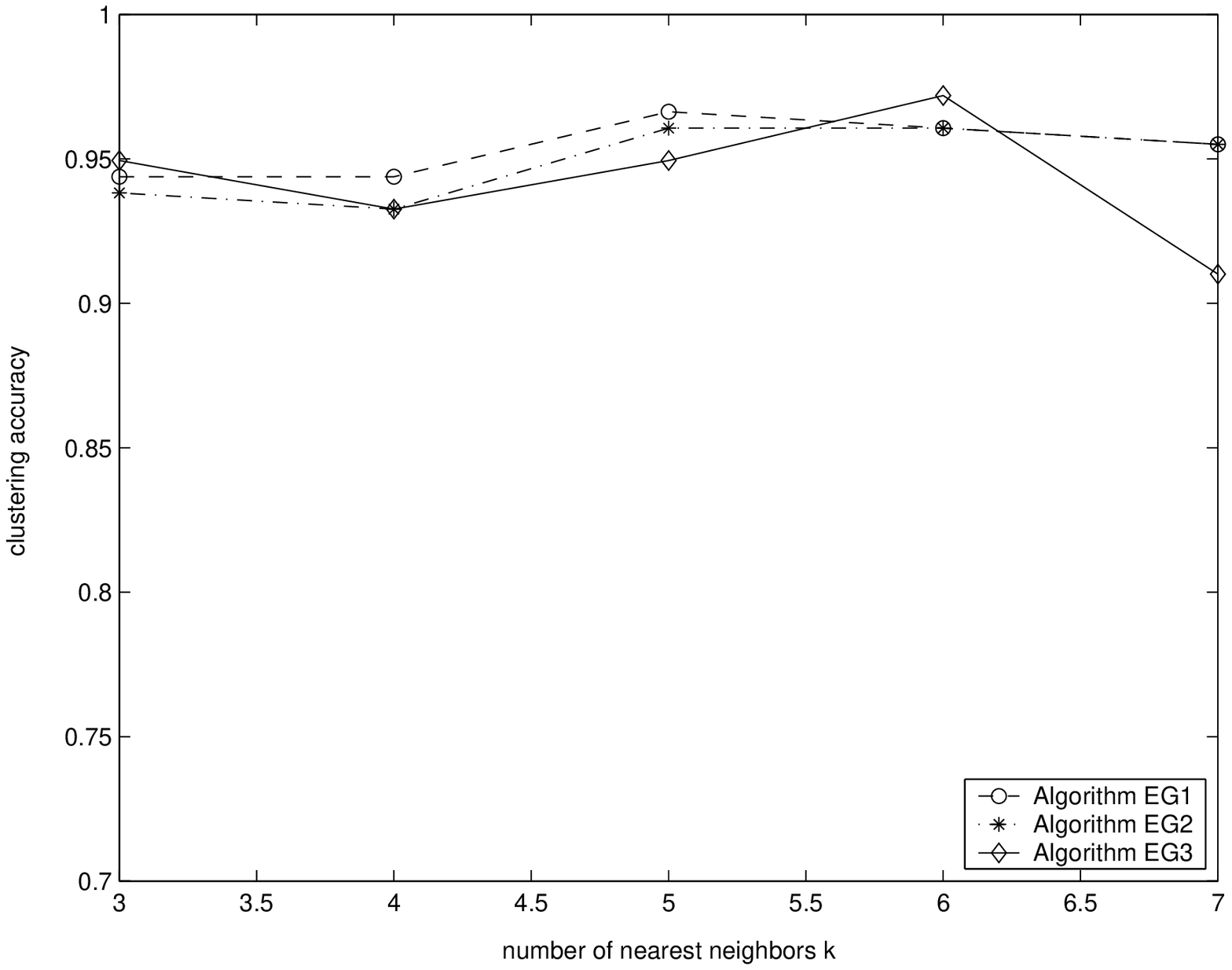}}\\
\vspace*{8pt} \centering \subfigure[Ionosphere dataset]{
\includegraphics[width=0.3\textwidth]{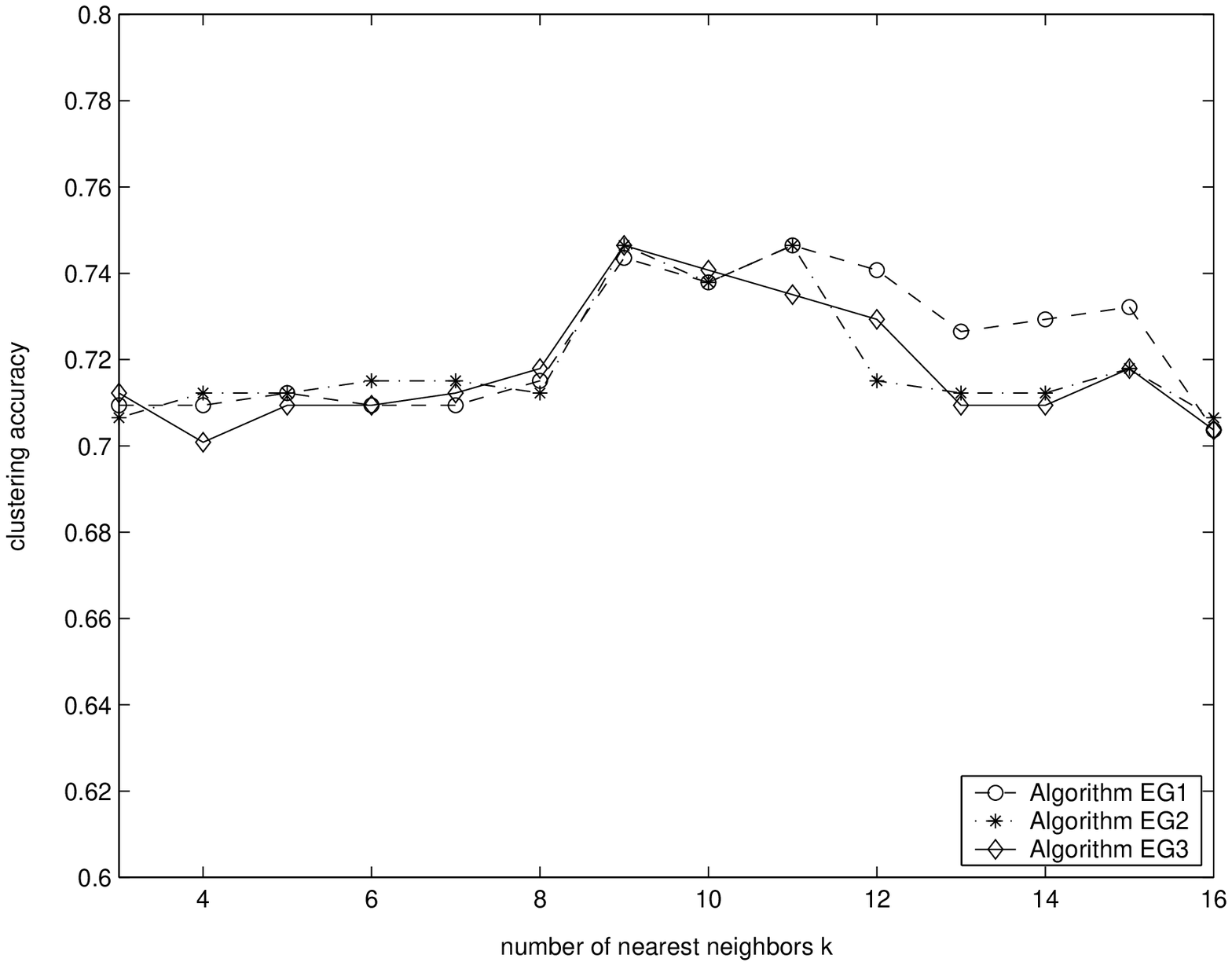}}
\subfigure[Breast dataset]{
\includegraphics[width=0.3\textwidth]
{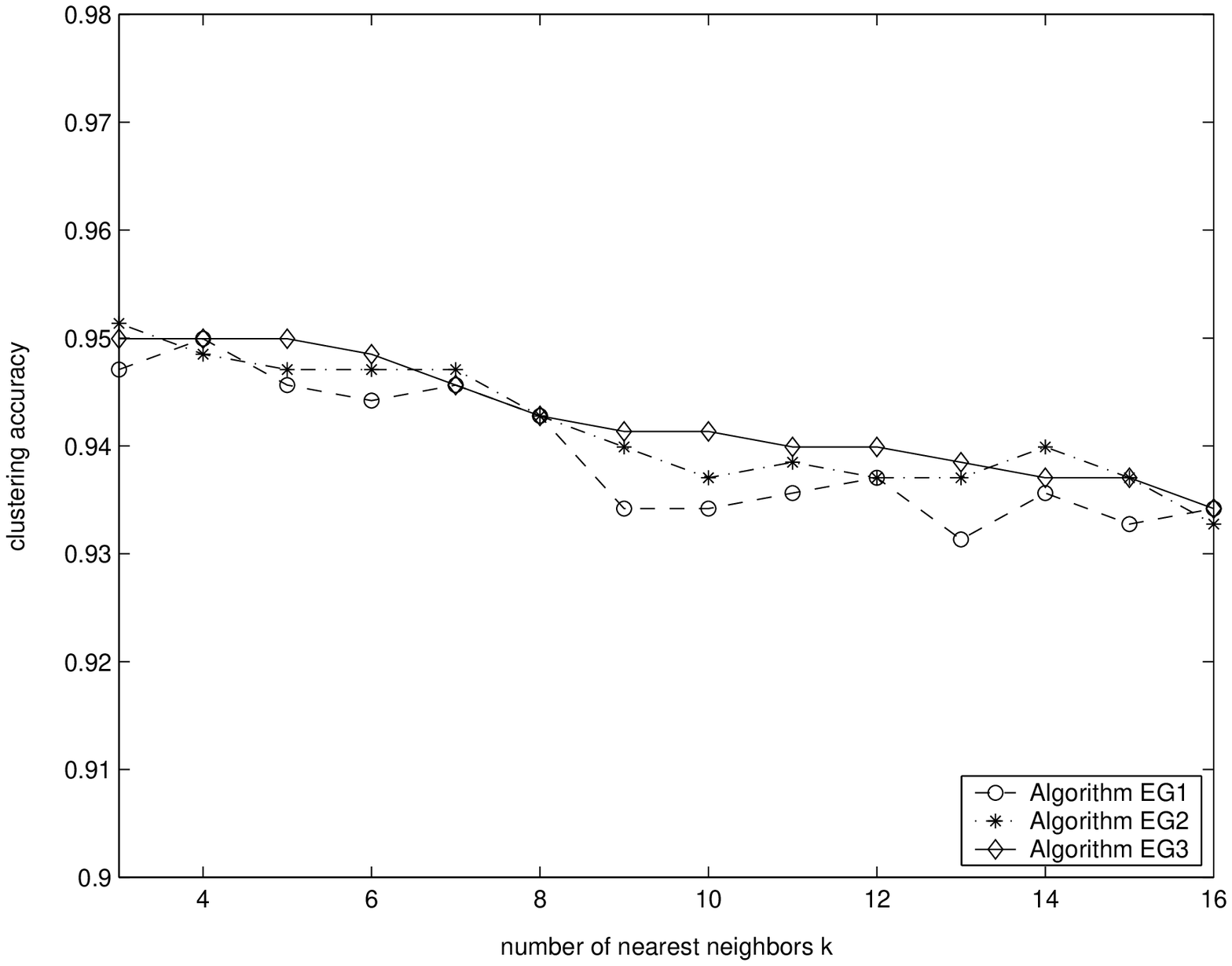}}\\
\caption{Comparison of clustering accuracies in Algorithm EG1, EG2,
and EG3.}\label{fig:5}
\end{figure}

Three clustering algorithms are applied on the five datasets
respectively. Because the capacity of exploration of an ERR function
depends in part on the number \textit{k} of nearest neighbors, the
algorithms are run on every dataset at different numbers of nearest
neighbors. Those clustering results obtained by three algorithms are
compared in Fig.~\ref{fig:5}, in which each point represents a
clustering accuracy. The clustering accuracy is defined as below:
\begin{definition}
$cst_{i}$ is the label which is assigned to a data point
$\textbf{\textit{X}}_{i}$ in a dataset by the algorithm, and $c_{i}$
is the actual label of the data point $\textbf{\textit{X}}_{i}$ in
the dataset. So the clustering accuracy is~\cite{Erkan2006}:
\begin{equation}
\begin{array}{ll}accuracy=\sum^{n}_{i=1}\lambda(map(cst_{i}), c_{i})/n\vspace{6pt}\\
\lambda(map(cst_{i}), c_{i})=\left\{\begin{array}{lll}1 & \textrm{ if $map(cst_{i})=c_{i}$}\vspace{4pt}\\
0 & \textrm{ otherwise}\end{array} \right.
\end{array}
\end{equation}
where the mapping function $map(\cdot)$ maps the label got by the
algorithm to the actual label.
\end{definition}


As is shown in Fig.~\ref{fig:5}, the clustering results obtained by
Algorithm EG1 and EG2 are similar at different numbers of nearest
neighbors. As a whole, the results of Algorithm EG1 are a bit better
than that of Algorithm EG2 owing to the stronger capacity of
exploration. However, for the same dataset the best result is
achieved by Algorithm EG3, which shows the strongest capacity of
exploration. As analyzed above, both the strongest capacity of
exploration and the stable results cannot be achieved at the same
time, so a trade-off is needed between them.

We compare our results to those results obtained by other clustering
algorithms, Kmeans~\cite{Ding2007}, PCA-Kmeans~\cite{Ding2007},
LDA-Km~\cite{Ding2007}, on the same dataset. The comparison is
summarized in Table~\ref{tab:3}.
\begin{table}[htbp]
\caption{Comparison of clustering accuracies of
algorithm.}\label{tab:3} \centering {\begin{tabular} {ccccccc}
\hline Algorithm & Soybean & Iris & Wine & Ionosphere & Breast
\\\hline
EG1 & \hphantom{0}95.75\% & \hphantom{0}90\% & \hphantom{0}96.63\% & \hphantom{0}74.64\% & \hphantom{0}94.99\% \\
EG2 & \hphantom{0}91.49\% & \hphantom{0}90.67\% & \hphantom{0}96.07\% & \hphantom{0}74.64\% & \hphantom{0}95.14\% \\
EG3 & \hphantom{0}97.87\% & \hphantom{0}90.67\% & \hphantom{0}97.19\% & \hphantom{0}74.64\% & \hphantom{0}94.99\% \\
Kmeans & \hphantom{0}68.1\% & \hphantom{0}89.3\% & \hphantom{0}70.2\% & \hphantom{0}71\% & \hphantom{0}-- \\
PCA-Kmeans & \hphantom{0}72.3\% & \hphantom{0}88.7\% & \hphantom{0}70.2\% & \hphantom{0}71\% & \hphantom{0}-- \\
LDA-Km & \hphantom{0}76.6\% & \hphantom{0}98\% & \hphantom{0}82.6\% & \hphantom{0}71.2\% & \hphantom{0}-- \\
\hline
\end{tabular}}
\end{table}

\subsection{Discussions}
In the subsection, firstly, we discuss how the number of clusters is
affected by the number \textit{k} of nearest neighbors changing.
Then, for the Algorithm EG1, the relationships between the ratio of
exploration and the clustering results is investigated, which
provides a way to select the ratio of exploration $\eta$. Finally,
the rates of convergence in three clustering algorithms are
compared.

\subsubsection{Number of nearest neighbors vs. number of clusters}
The number \textit{k} of nearest neighbors represents the number of
neighbors to which a data point
$\textbf{\textit{X}}_{i}\in\textbf{\textit{X}}$ connects. For a
dataset, the number \textit{k} of nearest neighbors determines the
number of clusters in part. Generally speaking, the number of
clusters decreases inversely with the number \textit{k} of
nearest neighbors. If the number \textit{k} of nearest neighbors is
small, which indicates a data point $\textbf{\textit{X}}_{i}$
connects to a few neighbors, in this case the area that the ERR
function may explore is also small, i.e., the elements in the union
of the extended neighbor set $\Upsilon_{t-1}(i)$ and the neighbor
set $\Gamma_{t-1}(i)$ are only a few. Therefore, when the network
evolves over time, strategies are spread only in a small area.
Finally many small clusters with evolutionarily stable strategies
appear in the network. On the other hand, a big number \textit{k} of
nearest neighbors provides more neighbors for a data point, which
implies that the cardinality of the union is larger than that when a
small \textit{k} is employed. This also means that a larger area can
be observed and explored by the ERR function, so that big clusters
containing more data points are formed because evolutionarily stable
strategies are spread in larger areas.

\begin{figure}[htbp]
\centering \subfigure[$ k=8$]{
\includegraphics[width=0.3\textwidth]{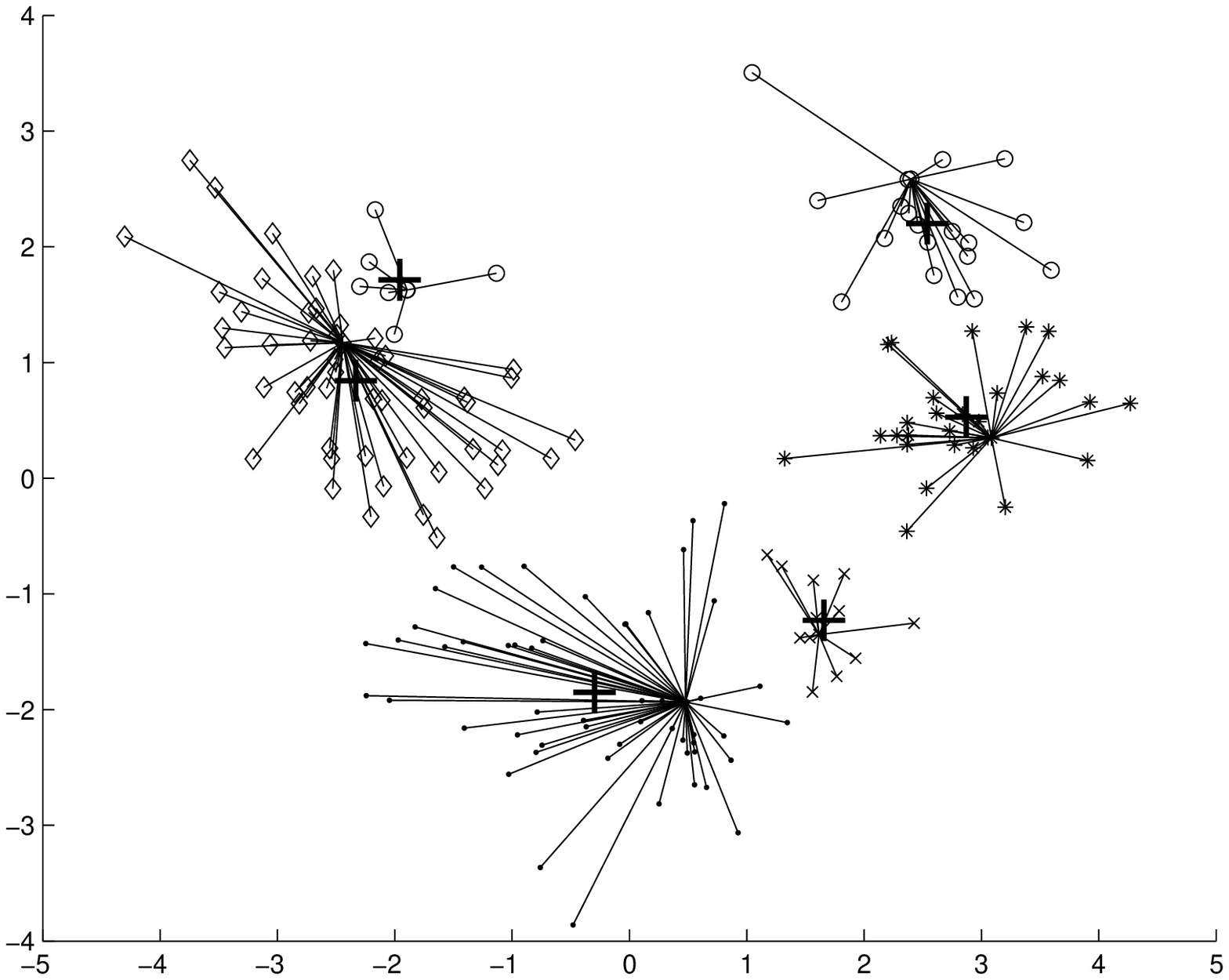}} \subfigure[$ k=10$]{
\includegraphics[width=0.3\textwidth]{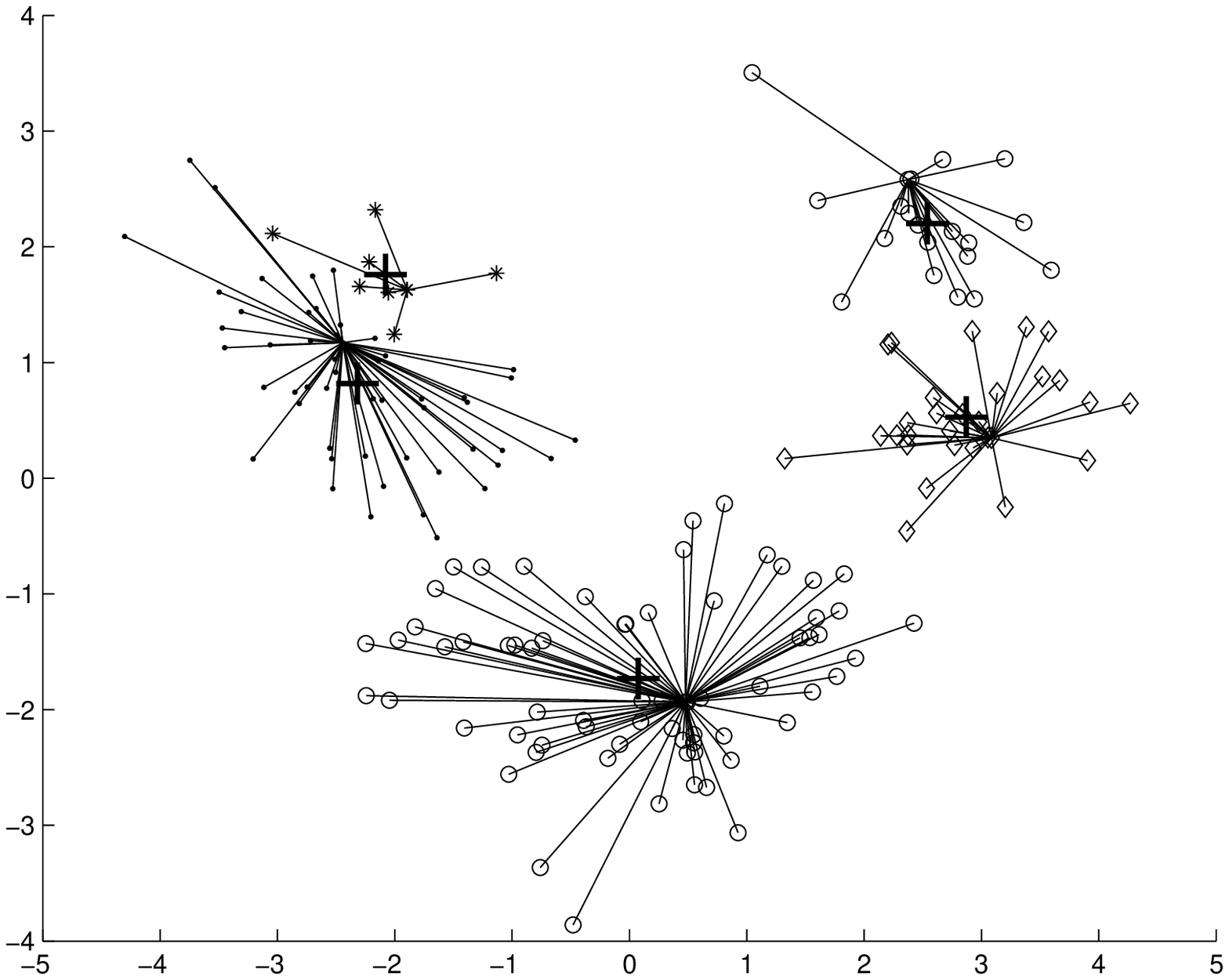}}
\subfigure[$
k=15$]{ \includegraphics[width=0.3\textwidth]{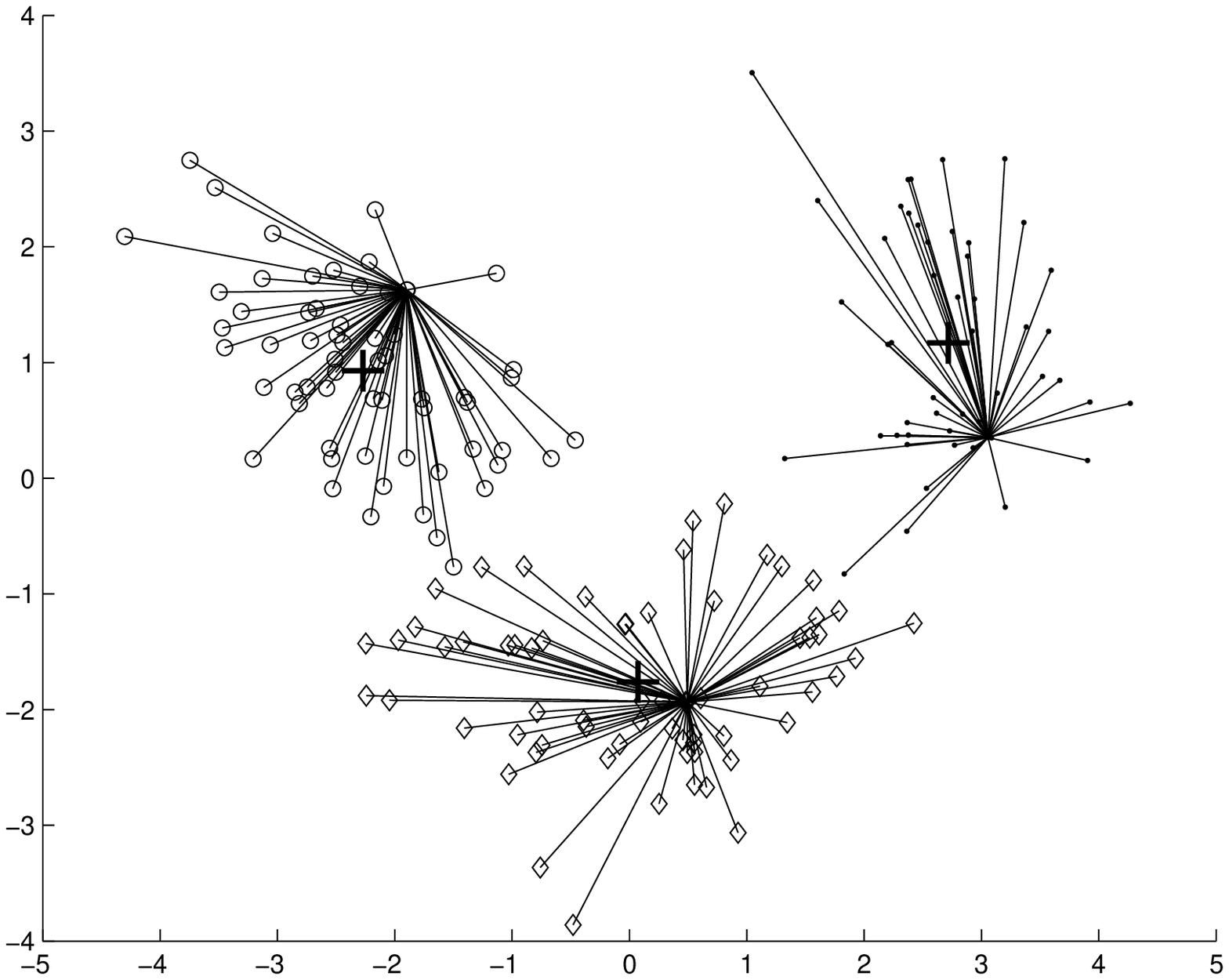}}\\
\caption{The number of nearest neighbors vs. number of
clusters.}\label{fig:1}
\end{figure}

For a dataset, the clustering results in different number \textit{k}
of nearest neighbors have been illustrated in Fig.~\ref{fig:1}, in
which each data point only connects to the neighbor with the largest
preference, and clusters are represented by different signs. As is
shown in Fig.~\ref{fig:1}, we can find that only a few data points
receive considerable connections, whereas most of data points have
only one connection. This indicates that when the evolution of
network is ended, the network formed is characterized by the
scale-free network \cite{Barabasi2003}, i.e., winner takes all.
Besides, in Fig.~\ref{fig:1}(a), six clusters are obtained by the
clustering algorithm, when $k=8$. As the number \textit{k} of
nearest neighbors rises, five clusters are obtained when $k=10$,
three clusters when $k=15$. So, if the exact number of clusters is
not known in advance, different number of clusters may be achieved
by adjusting the number of nearest neighbors in practice.

\subsubsection{Effect of the ratio of exploration in Algorithm EG1}
\begin{figure}[htbp]
\centering
\includegraphics[width=0.5\textwidth]{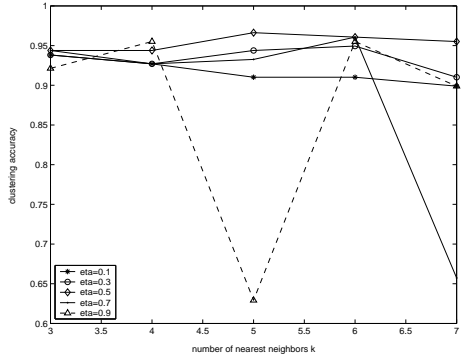}
\caption{The results of algorithm EG1 at different rates of
exploration.}\label{fig:2}
\end{figure}

For the ERR function $B^{1}_{i}(\cdot)$ used in Algorithm EG1, its
capacity of exploration may be adjusted by setting different ratio
of exploration $\eta\in[0,1]$. If the ratio of exploration is
$\eta=0$, then the extended neighbor set formed will be empty.
Hence, the network does not evolve over time, since the ERR function
does not explore. On the other hand, when the ratio of exploration
takes the maximum $\eta=1$, the ERR function is with the strongest
capacity of exploration, because an extended neighbor set formed by
all neighbors can be observed. Then we may ask naturally: what
should the ratio of exploration be taken? To answer this question,
we compare those clustering results at different $\eta$ shown in
Fig.~\ref{fig:2}, in which the results are represented by the
clustering accuracies.

From Fig.~\ref{fig:2}, we can see that when the ratio of exploration
is greater than 0.5, the clustering results obtained by Algorithms
EG1 fluctuates strongly, as seems over exploration. On the contrary,
in the case when $\eta\leq0.5$, the clustering results are stable
relatively. As a whole, when the ratio of exploration $\eta=0.5$,
the best results are achieved in Algorithm EG1. In conclusion, a big
ratio of exploration is not too good, because it may cause the ERR
function $B^{1}_{i}(\cdot)$ over exploration. However, if the ratio
of exploration is too small, good results are not obtained because
of a lack of exploration. Therefore, in the later discussion, the
ratio of exploration in Algorithm EG1 takes $\eta=0.5$.

\subsubsection{Three ERR functions vs. rate of convergence} 
\begin{figure}[htbp]
\centering \subfigure[EG1]{
\includegraphics[width=0.3\textwidth]{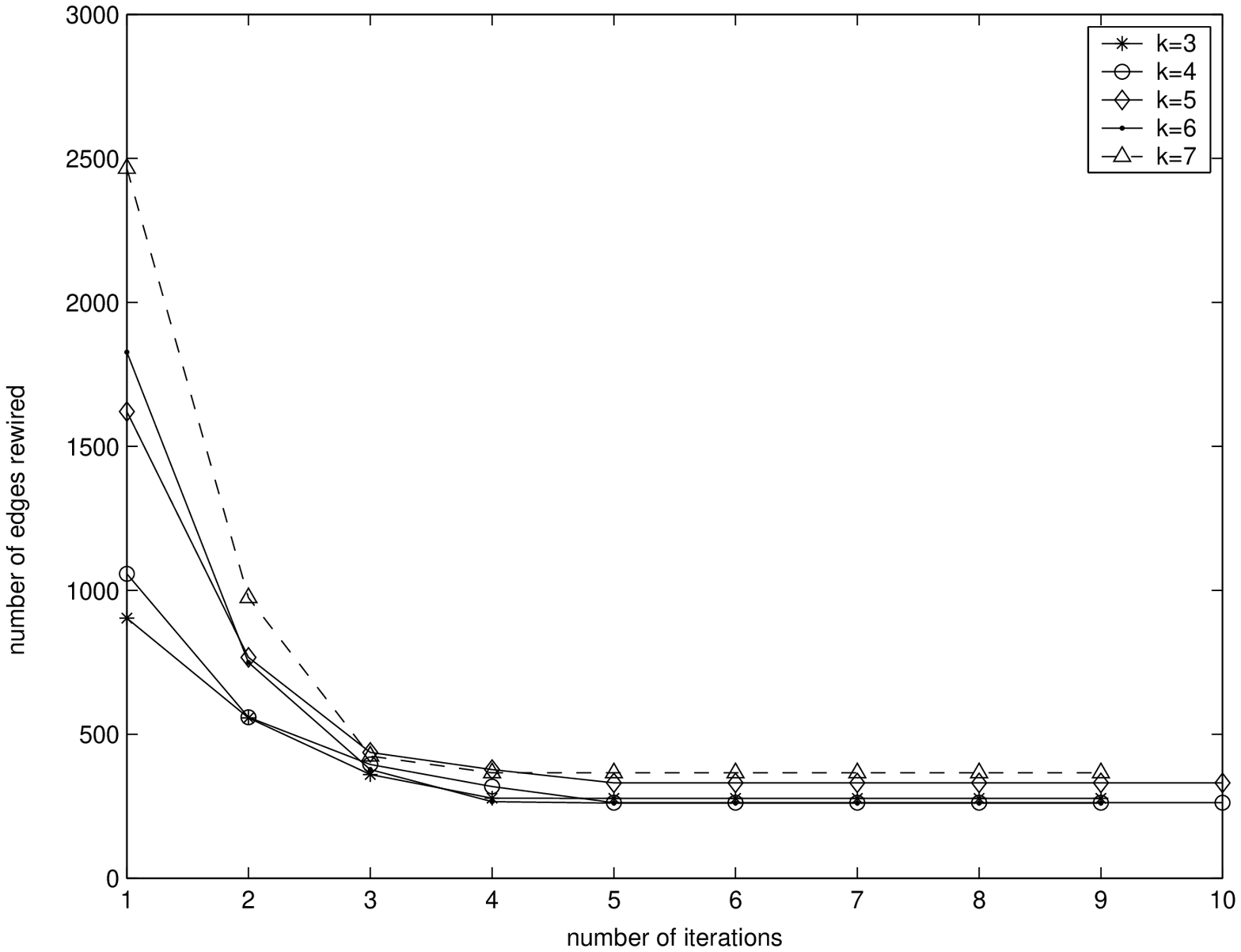}}
\subfigure[EG2]{
\includegraphics[width=0.3\textwidth]
{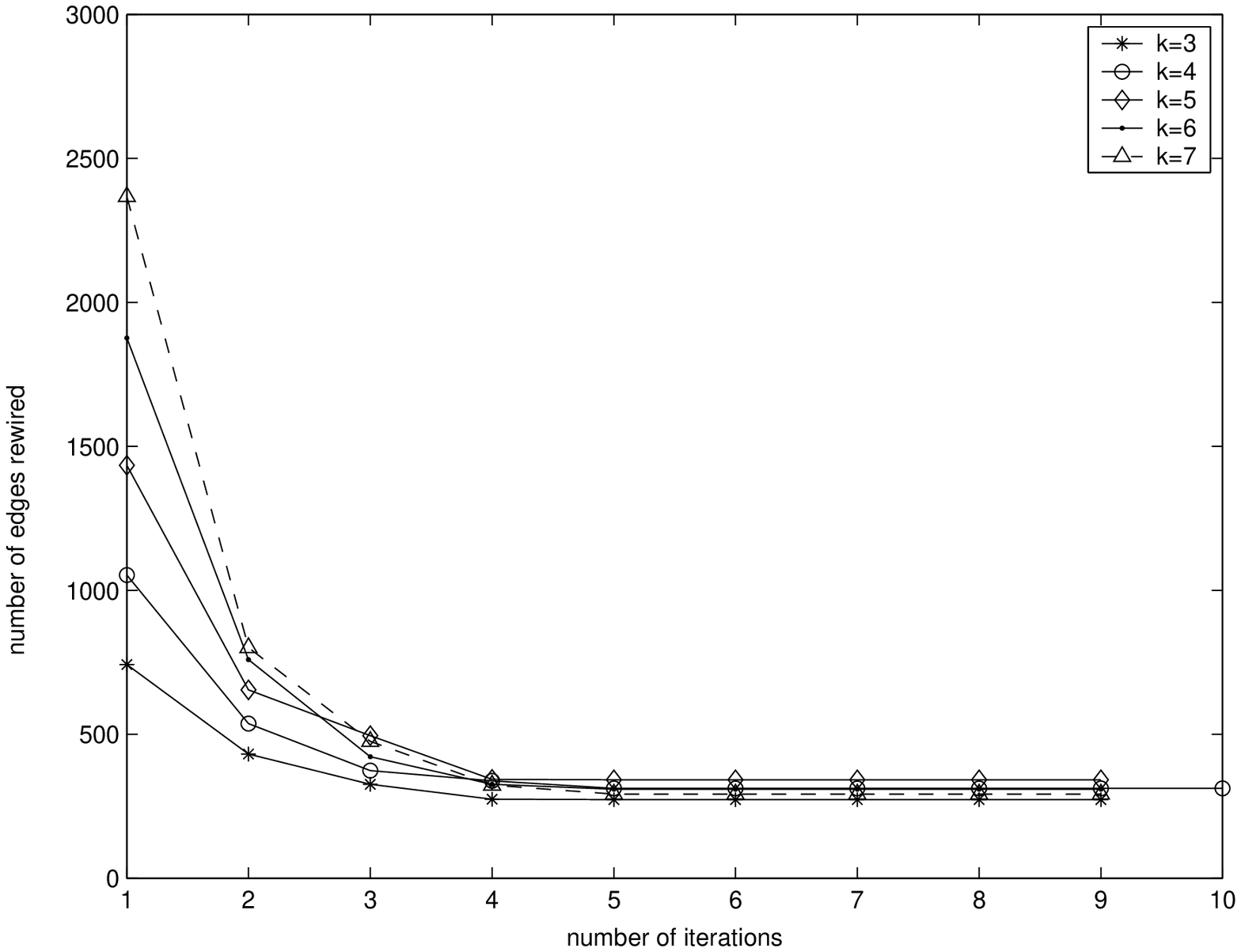}}
\subfigure[EG3]{ \includegraphics[width=0.3\textwidth]{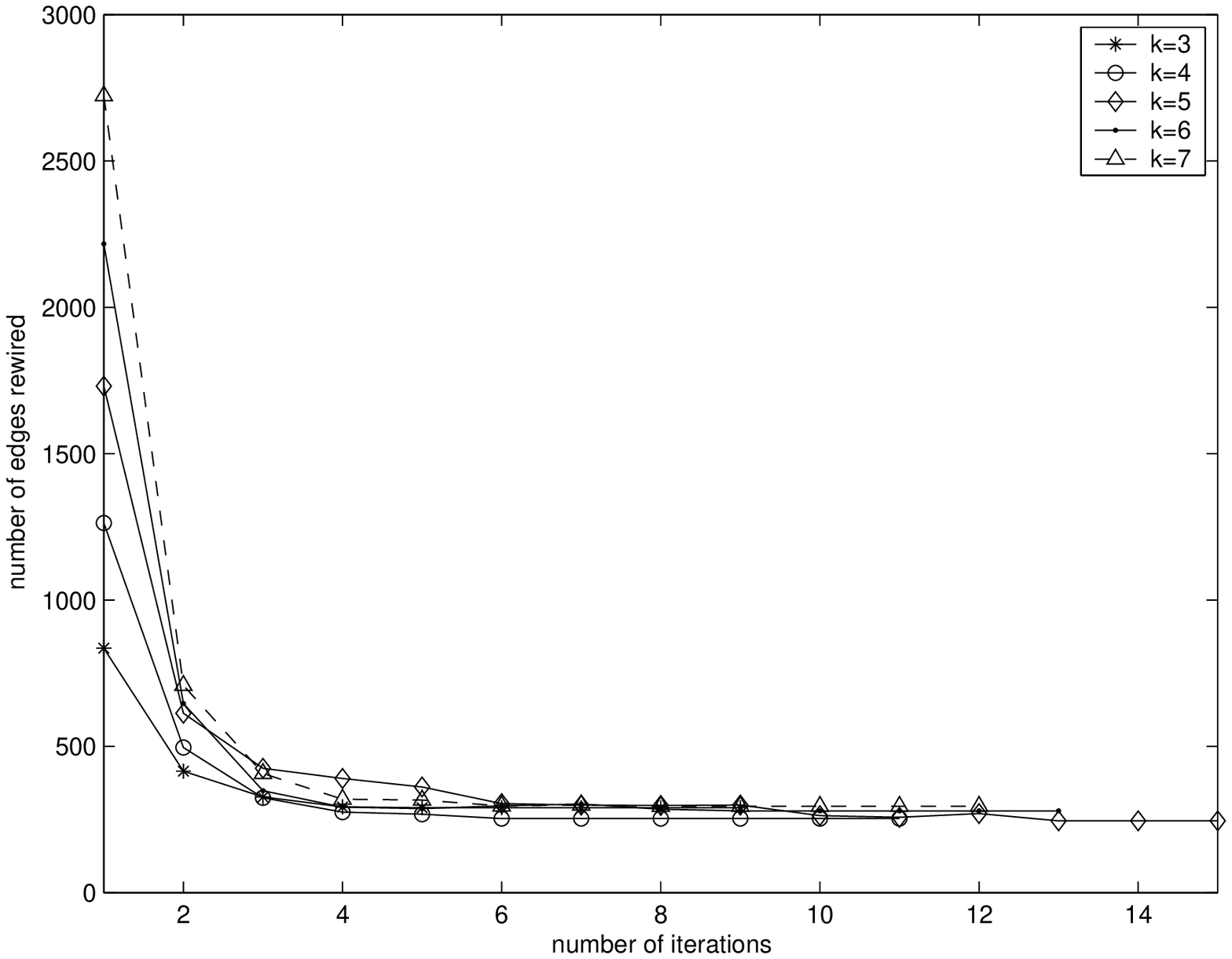}}\\
\caption{The number of edges rewired or rates of convergence of
three algorithms.}\label{fig:3}
\end{figure}

As for
Algorithm EG1, when $\eta=0.5$, the ERR function $B^{1}_{i}(\cdot)$
may observe the extended neighbor set formed by half of neighbors.
In this case, the payoff threshold is
$\theta_{t-1}^{1}(i)=median(\{u_{t-1}(j),j\in\Gamma_{t-1}(i)\})$.
In Algorithm EG2, however, the ERR function $B^{2}_{i}(\cdot)$ only
can observe the extended neighbor set formed by neighbors whose
payoffs are greater than the average $\theta_{t-1}^{2}(i)$.
Generally speaking, for the same \textit{k}, the median of payoffs
is smaller than or equal to the mean, i.e.,
$\theta_{t-1}^{1}(i)\leq\theta_{t-1}^{2}(i)$. So the exploring area
of the ERR function $B^{1}_{i}(\cdot)$ is larger than that of the
ERR function $B^{2}_{i}(\cdot)$, that is, the number of edges
rewired in Algorithm EG1 is more than that in Algorithm EG2. In
addition, the number of edges rewired in Algorithm EG3 is largest,
since in Algorithm EG3 the ERR function $B^{3}_{i}(\cdot)$ provides
stronger capacity of exploration for players with small payoffs,
which makes more edges are removed and rewired. The comparison of
number of edges rewired in three algorithms is illustrated in
Fig.~\ref{fig:3}. As is shown in Fig.~\ref{fig:3}, at different
number of nearest neighbors, the number of edges rewired in
Algorithm EG3 is larger than that in other two, and the number of
edges rewired in Algorithm EG1 is larger than Algorithm EG2. For
each one of three algorithms, the number of edges rewired increases rapidly with the number of nearest neighbors.

Besides, the number of iterations indicates the rate of convergence
of an algorithm. From Fig.~\ref{fig:3}, we can see that the rates of
convergence in Algorithm EG1 and EG2 are almost the same, and the
rate of Algorithm EG3 is slower slightly than the other two because
the ERR function $B^{3}_{i}(\cdot)$ in Algorithm EG3 explores larger
areas than that in Algorithm EG1 and EG2. For each one of three algorithms,
as the number of nearest neighbors rises, the number of iterations
also rises slightly.

\begin{figure}[htbp]
\centering \subfigure[EG1]{
\includegraphics[width=0.3\textwidth]{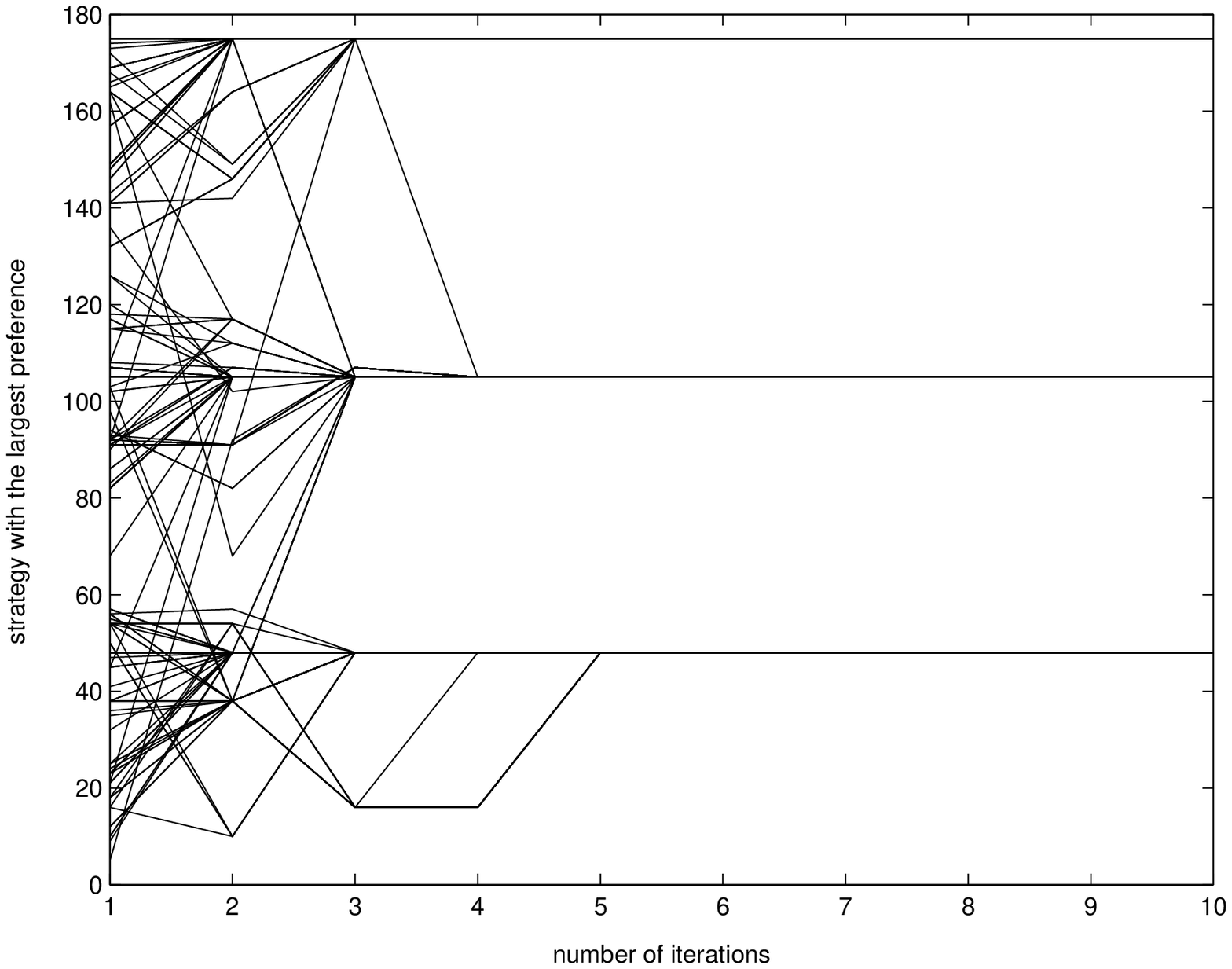}}
\subfigure[EG2]{
\includegraphics[width=0.3\textwidth]
{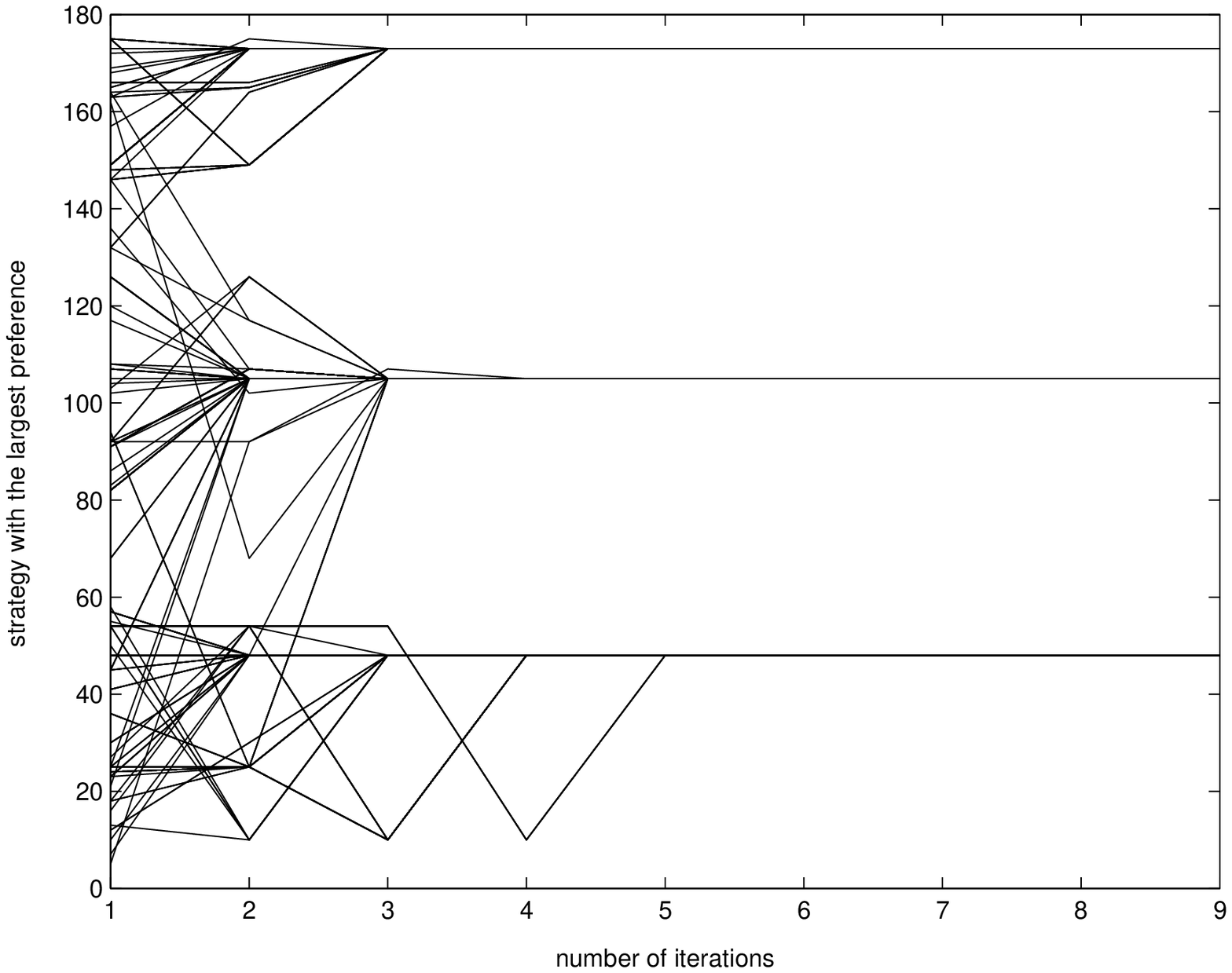}}
\subfigure[EG3]{ \includegraphics[width=0.3\textwidth]{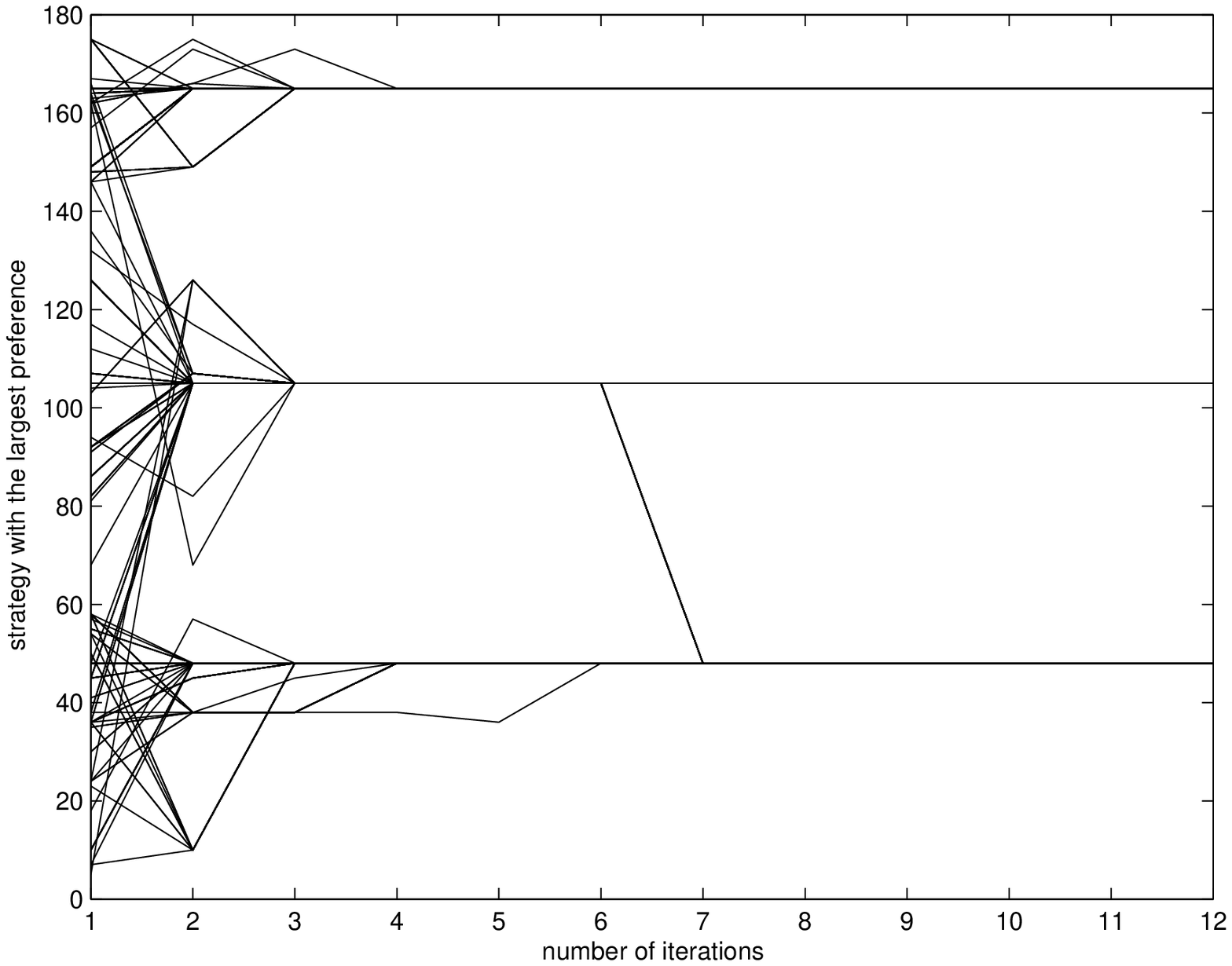}}\\
\caption{Evolutions of strategies with the largest preference and
evolutionarily stable strategies.}\label{fig:4}
\end{figure}

When the algorithm converges, the evolutionarily stable strategies
appear in the network at the same time. The changes of the
strategies with the largest preference for each data point are shown
in Fig.~\ref{fig:4}, where the straight lines in the right side of
figures represent the evolutionarily stable strategies, and the
number of straight lines is the number of clusters. As is shown in
Fig.~\ref{fig:4}, the evolutionarily stable strategies appear in the
network after only a few iterations, as indicates the rates of
convergence of algorithms are fast enough.

\section{Conclusion}
A model based upon games on an evolving network has been
established, which may be used to explain the formation of groups in
society partly. Following this model, three clustering algorithms
(EG1, EG2 and EG3) using three different ERR functions are
constructed, in which each data point in a dataset is regarded as a
player in a game. When a distance function is selected, the initial
network is created among data points according to Definition 1. Then, by
applying an ERR function, the network will evolve over time due to
edges removed and rewired. Hence, the preference set of a player needs to be
adjusted in terms of Definition 5, and payoffs of players are recomputed
too. During the network evolving, certain strategies are spread in
the network. Finally, the evolutionarily stable strategies emerge in
the network. According to evolutionarily stable strategies played by
players, those data points with the same evolutionarily stable
strategies are collected as a cluster. As such, the clustering
results are obtained, where the number of evolutionarily stable
strategies corresponds to the number of clusters.

The ERR functions
($B^{1}_{i}(\cdot),B^{2}_{i}(\cdot),B^{3}_{i}(\cdot)$) employed in
three clustering algorithms provide different capacities of
exploration for these clustering algorithms, i.e., the sizes of
areas which they can observe are various. So, the clustering results
of three algorithms are different. For the ERR function
$B^{1}_{i}(\cdot)$, it is with a constant ratio of exploration
$\eta=0.5$ because over exploration may occur when $\eta>0.5$, and
can observe an extend neighbor set formed by half of neighbors. The
ERR function $B^{2}_{i}(\cdot)$, however, can observe an extend
neighbor set formed by neighbors whose payoffs are larger than the
average, while the ERR function $B^{3}_{i}(\cdot)$ provides stronger
capacity of exploration for data points with small payoffs. Besides,
the clustering results of Algorithm EG1 and EG2 are more stable than
that of Algorithm EG3, but the best results are achieved by
Algorithm EG3 due to the strongest capacity of exploration among
three algorithms.

In the case when the exact number of clusters is unknown in advance,
one can adjust the number \textit{k} of nearest neighbors to control
the number of clusters, where the number of clusters decreases inversely with the number \textit{k} of nearest neighbors. We
evaluate the clustering algorithms on five real datasets,
experimental results have demonstrated that data points in a dataset
are clustered reasonably and efficiently, and the rates of
convergence of three algorithms are fast enough.

\bibliographystyle{hieeetr}
\bibliography{manuscript}

\end{document}